\documentclass{article}
\usepackage{spconf,amsmath,graphicx}
\usepackage{epstopdf}
\usepackage{subfig}
\usepackage{amssymb}
\usepackage{amsmath}
\usepackage{longtable}
\usepackage{array}
\usepackage{multirow}
\usepackage{url}
\usepackage{algorithm,color,cite}
\usepackage{algpseudocode}
\usepackage{threeparttable}

\def\x{{\mathbf x}}

\def\W{{\mathbf W}}

\def\D{{\mathbf D}}

\def\C{{\mathbf C}}

\def\x{{\mathbf x}}

\title{DEEP TOPIC MODELING By MULTILAYER BOOTSTRAP NETWORK and LASSO}
\name{Jianyu Wang$^2$ and Xiao-Lei Zhang$^{1,2}$}
\address{$^1$Research \& Development Institute of Northwestern Polytechnical University in Shenzhen, China \\
$^2$Center for Intelligent Acoustics and Immersive Communications and\\
  School of Marine Science and Technology, Northwestern Polytechnical University, China\\
\thanks{Email: alexwang96@mail.nwpu.edu.cn, xiaolei.zhang@nwpu.edu.cn}
\thanks{
This work was supported in part by the Project of the Science, Technology, and Innovation Commission of Shenzhen Municipality under grant No. JCYJ20170815161820095, in part by the National Natural Science Foundation of China (NSFC) funding scheme under Project No. 61671381, and in part by the Shaanxi Natural Science Basic Research Program under grant No. 2018JM6035.}}
\begin{document}
\ninept
\maketitle
\begin{abstract}
Topic modeling is widely studied for the dimension reduction and analysis of documents. However, it is formulated as a difficult optimization problem. Current approximate solutions also suffer from inaccurate model- or data-assumptions. To deal with the above problems, we propose a polynomial-time deep topic model with no model and data assumptions. Specifically, we first apply multilayer bootstrap network (MBN), which is an unsupervised deep model, to reduce the dimension of documents, and then use the low-dimensional data representations or their clustering results as the target of supervised Lasso for topic word discovery. To our knowledge, this is the first time that MBN and Lasso are applied to unsupervised topic modeling. Experimental comparison results with five representative topic models on the 20-newsgroups and TDT2 corpora illustrate the effectiveness of the proposed algorithm.
\end{abstract}

%
\section{Introduction}
\label{sec:intro}
Topic modeling is an unsupervised method that learns latent structures and salient features from document collections. It is originally formulated as a hierarchical generative model: a document is generated from a mixture of topics, and a word in the document is generated by first choosing a topic from a document-specific distribution, and then choosing the word from the topic-specific distribution. The main difficulty of topic modeling is the optimization problem, which is NP-hard in the worst case due to the intractability of the posterior inference. Existing methods aim to find approximate solutions to the difficult optimization problem, which falls into the framework of matrix factorization.

Matrix factorization based topic modeling maps documents into a low-dimensional semantic space by decomposing the documents into a weighted combination of a set of topic distributions: $\mathbf{D} \approx \mathbf{C} \mathbf{W}$ where $\mathbf{D}(:,d)$ represents the $d$-th document which is a column vector over a set of words with a vocabulary size of $v$, $\C(:, g)$ denotes the $g$-th topic which is a probability mass function
over the vocabulary, and $W(g, d)$ denotes the probability
of the $g$-th topic in the $d$-th document. Existing methods for the matrix decomposition can be categorized to two classes in general---probabilistic models \cite{blei2003latent,iwata2013unsupervised,iwata2018topic,azarbonyad-hitr-2018} and nonnegative matrix factorizations (NMF) \cite{gillis2014fast,xuan2018doubly,Bobadilla2018RecommenderSC}.

A seminal work of probabilistic models is latent Dirichlet allocation (LDA) \cite{blei2003latent}.
It assumes that each document is a sample from a multinomial distribution whose parameters are generated from $\mathbf{C}\mathbf{W}(:, d)$. Each column of $\mathbf{C}$ and $\mathbf{W}$
also represent multinomial distributions independently
drawn from Dirichlet distributions. It adopts Kullback-Leibler divergence to measure the distance between $\mathbf{D}$ and $\mathbf{C}\mathbf{W}$, since the posterior distributions $p(\mathbf{W}|\mathbf{D})$ and $p(\mathbf{C}|\mathbf{D})$  are coupled. Later on, many models followed the above framework, such as hierarchical Dirichlet process \cite{teh2005sharing,li2018supervised} and Laplacian probabilistic semantic indexing \cite{cai2008modeling}. However, the model assumptions, such as the multinomial distribution, may not be always accurate.

The validness of NMF comes from the fact that the matrices $\mathbf{C}$ and $\mathbf{W}$ should be nonnegative.
The objective function of NMF is generally as follows:
\begin{equation}\label{eq2}
  (\mathbf{C},\mathbf{W}) = \arg \min_{\mathbf{C}\geq \mathbf{0};\mathbf{W}\geq \mathbf{0}}\parallel{\mathbf{D}-\mathbf{C}\mathbf{W}}\parallel^{2}_{F}
\end{equation}
An important weakness of this formulation is that there is no guarantee that the solutions of $\mathbf{C}$ and $\mathbf{W}$ are unique \cite{xu2003document}. To solve the identifiability problem, many NMF methods adopted an \textit{anchor word} assumption, which assumes that every topic
has a characteristic anchor word that does not appear in the other topics \cite{huang2016anchor}. However, this assumption may not always hold in practice. Recently, an anchor-free NMF based on the second-order statistics of documents \cite{fu2018anchor} has been proposed, which significantly improved the performance of NMF methods. Another problem of NMF is that it is formulated as a shallow learning method, which may not capture the nonlinearity of documents.


Motivated by the above problems, this paper proposes a deep topic model (DTM), which learns a deep representation of the documents, i.e. $f(\mathbf{D})$, and the topic-word matrix $\C$ separately, under the assumption that if each of the components is good enough, then the overall performance can be boosted. Specifically, we apply multilayer bootstrap networks (MBN) \cite{zhang2018multilayer} to learn a document-topic projection $f(\cdot)$ first, and then apply Lasso \cite{Tibshirani94regressionshrinkage} to learn $\C$ given $f(\mathbf{D})$. MBN is a simple nonparametric deep model for unsupervised dimension reduction, which overcomes the problems of model assumptions, shallow learning, and anchor word assumption. Given the output of MBN, the topic modeling becomes a simple supervised regression problem. We employ Lasso for this problem. Empirical  results on the TDT2 and 20-newsgroups corpora illustrate the effectiveness of the proposed algorithm.

\section{Deep topic modeling}
\label{sec:pagestyle}

\subsection{Object function}
\begin{figure}[!htb]
  \centering
  \includegraphics[scale = 0.6]{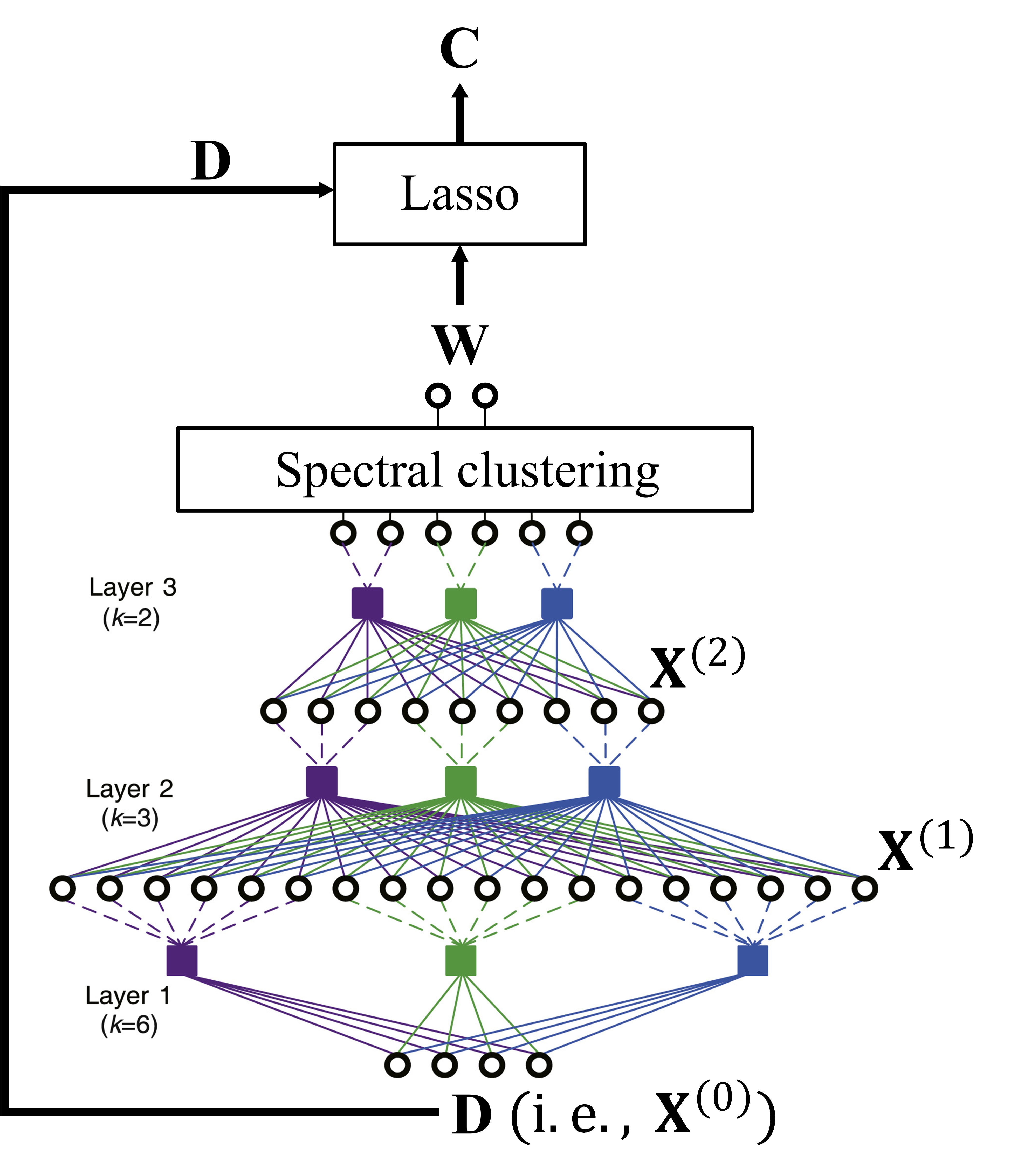}\\
  \caption{Deep topic model.}\label{fig1}
\end{figure}

The objective of DTM is defined as
\begin{equation}\label{eq:dtm}
  \min_{f(\cdot), \mathbf{C}}   \frac{1}{2}\| \mathbf{C}f(\D)-\mathbf{D}\|_F^2 + \lambda\Omega(\mathbf{C})
\end{equation}
where $f(\cdot)$ is an unsupervised deep model containing multiple layers of nonlinear transforms, $\Omega(\cdot)$ is a regularizer, and $\lambda$ is a regularization hyperparameter. We optimize \eqref{eq:dtm} in two steps. The first step learns $f(\D)$ by MBN, which outputs the document-topic matrix $\W$. The second step learns the topic-word matrix $\C$ by Lasso, given $\W=f(\D)$. The overall DTM algorithm is shown in Fig. \ref{fig1}.

\subsection{Multilayer bootstrap network}

The network structure of MBN is shown in Fig. \ref{fig1}. It is a deep dimensionality reduction algorithm optimized by random resampling of data and one-nearest-neighbor optimization \cite{zhang2018multilayer}. It consists of $L$ gradually narrowed hidden layers from bottom-up. Each hidden layer consists of $V$ $k$-centroids clusterings ($V>100$), where parameter $k$ at the $l$-th layer is denoted by $k_l$, $l=1,\ldots,L$. Each $k_l$-centroids clustering has $k_l$ output units, each of which indicates one cluster. The output layer is the linear-kernel-based spectral clustering \cite{NIPS2001_2092}. It outputs $\W$, which is used as the input of the Lasso component.

MBN is trained layer-by-layer from bottom-up. To train the $l$-th layer, we simply need to focus on training each $k_l$-centroids clustering as follows:
\begin{itemize}
  \item \textbf{Random sampling of input}. The first step randomly selects $k_l$ data points from $\mathbf{X}^{(l-1)} = [\x_{1}^{(l-1)},\ldots,\x_{N}^{(l-1)}]$ as the $k_l$ centroids of the clustering, where $N$ is the size of the corpus. If $l=1$, then $\mathbf{X}^{(l-1)}=\D$.
  \item \textbf{One-nearest-neighbor learning}. The second step assigns any input $\mathbf{x}^{(l-1)}$ to one of the $k_l$ clusters and outputs a $k_l$-dimensional indicator vector $\textbf{h}=[h_1,\dots,h_{k_l}]^T$, which is a one-hot sparse vector.
\end{itemize}
The output units of all $k_l$-centroids clusterings are concatenated as the input of their upper layer, i.e. $\x^{(l)} = [\textbf{h}_1^T,\ldots,\textbf{h}_V^T]^T$. From the above description, we can see that MBN does not make any model or data assumptions.

 Note that the parameter setting $\{k_l\}_{l=1}^L$ is important to maintain the tree property of MBN. In practice, it obeys the following criterion:
 \begin{eqnarray}
&& k_1 = \left\lfloor N/2 \right\rfloor,\quad k_l = \left\lfloor \delta k_{l-1} \right\rfloor\\
&& k_L \approx\left\{ \begin{array}{ll}
\lceil \frac{N_{Z}}{N_{z}}\rceil, &\mbox{if $\mathbf{D}$ is strongly class imbalanced}\\
 1.5c,& \mbox{otherwise}\\

\end{array}\right.\label{eq:xx}
 \end{eqnarray}
 where $\delta\in(0,1)$ is a user defined hyperparameter with $0.5$ as its default, $c$ is the number of topics, $N_Z$ and $N_z$ are the numbers of the documents belonging to the largest and smallest topics respectively. $\delta$ controls the network structure. \eqref{eq:xx} guarantees that at least one data point is sampled from each of the topics in probability. In other words, it ensures that the random samples at the top hidden layer is an effective model.

\subsection{Lasso}
Substituting the output of MBN, i.e. $\W$, to \eqref{eq:dtm} derives:
\begin{equation}\label{eq:lasso1}
\min_{ \mathbf{C}}   \frac{1}{2}\| \mathbf{C}\mathbf{W}-\mathbf{D}\|_F^2 + \lambda\Omega(\mathbf{C})
\end{equation}
\eqref{eq:lasso1} is a typical regularized regression problem \cite{pmlr-v54-yamada17a}. Many regression models can be applied to \eqref{eq:lasso1}. Here we choose Lasso, given its strong ability on the feature selection and prediction problems for high-dimensional data. Specifically, we use Lasso to calculate the conditional probability distribution of each word over the topics \cite{Tibshirani94regressionshrinkage}, which is formulated as the follow problem:
\begin{equation}\label{eq3}
  \min_{\mathbf{C}(i,:)}  \frac{1}{2}\parallel \mathbf{C}(i,:)\mathbf{W}-\mathbf{D}(i,:)\parallel_2^2 + \lambda\left\| \mathbf{C}(i,:)\right\|_1
\end{equation}
where $i=1,\ldots,v$ is the index of the $i$-th word. We adopt the \textit{alternating direction method of multipliers} (ADMM) \cite{Boyd10distributedoptimization} solver to solve problem \eqref{eq3}.\footnote{\url{https://github.com/foges/pogs}}

\section{Related work}
\label{sec:RelatedWork}

It is known that the main difficulty of hierarchical probabilistic topic models is the high computation on the inference problem of the hidden variables. Topic models based on deep variational auto-encoders overcome the difficulty. They generally
 can be decomposed into two modules:
an inference network $q(\mathbf{h}|\mathbf{D}(:,d))$ which compresses the documents into continuous hidden vectors $\mathbf{h}$ by deep neural networks, and a generative model $p(\mathbf{D}(:,d)|\mathbf{h})=\prod_{v=1}^V p({\mathbf{D}}(v,d)|\mathbf{h})$ which reconstructs the documents by generating the words independently from $\mathbf{h}$ \cite{miao2016neural} via restricted Boltzmann machines, sigmoid belief networks, Dirichlet processes, etc \cite{henao2015deep}, where $\mathbf{D}(v,d)$ is the $v$th word of the document $\mathbf{D}(:,d)$.
They maximise the evidence lower-bound of the joint likelihood of the documents and hidden variables:
\begin{equation}\label{eq:dntm}
  \mathcal{L} = \mathbb{E}_{q(\mathbf{h}|\mathbf{D}(:,d))} \left[ \sum_{v=1}^V \log p(\mathbf{D}(v,d)|\mathbf{h})\right] - D_{KL}[q(\mathbf{h}|\mathbf{D}(:,d))\Arrowvert p(\mathbf{h})]
\end{equation}
where $D_{KL}(\cdot\Arrowvert\cdot)$ denotes the Kullback-Leibler divergence between two distributions, $\mathbb{E}_{q(\mathbf{h}|\mathbf{D}(:,d))}(\cdot)$ is the expectation operator over $q(\mathbf{h}|\mathbf{D}(:,d))$, and $p(\mathbf{h})$ is a prior for $\mathbf{h}$.
The above models integrate the power of neural networks into the inference of the probabilistic topic models, which not only helps the probabilistic topic models scalable to big datasets but also speeds up the convergence of the probabilistic topic models significantly.
However, the prior assumption of $\mathbf{h}$ may not always hold, and moreover, the inference network
faces a problem of component collapsing \cite{srivastava2017autoencoding} which is a kind of bad local optima that is particularly endemic to auto-encoding variational Bayes and similar methods. On the contrary, the proposed method not only is able to generate deep representations of the documents but also does not suffer the aforementioned weaknesses.

\section{Experiments}
\label{sec:typestyle}

\subsection{Experimental settings}
   We conducted experiments on the 20-newsrgoups and the top 30 largest topics of the NIST Topic Detection and Tracking (TDT2) corpora. 20-Newsgroups consists of 18,846 documents with a vocabulary size of 26,214. The subset of TDT2 consists of 9,394 documents with a vocabulary size is 36,771. For each corpus, we randomly sampled $c=5,10,15,20$ topics respectively, and reported the average results over 50 Monte-Carlo runs. The indices of the 50 runs on TDT2 are the same as those at \url{http://www.cad.zju.edu.cn/home/dengcai/Data/TextData.html}. We used TF-IDF as the feature. We used cosine similarity to measure the similarity of two documents in the TF-IDF space.

For the proposed DTM, we set the hyperparameters of MBN and Lasso to their default values, i.e. $V=400$, $\delta=0.5$, and $\lambda=1/3$.\footnote{The default value of $\lambda$ is in the implementation of the ADMM algorithm.} We compared DTM to the following five representative topic modeling methods:
\begin{itemize}
\itemsep=0.0pt
\item \textbf{LDA}\cite{blei2003latent}: it is a seminal probabilistic model based on multinomial and Dirichlet distributions.
\item  \textbf{Locally-consistent topic modeling (LTM)} \cite{Cai:2009:PDD:1553374.1553388}: it extends the probabilistic latent semantic indexing algorithm \cite{hofmann2017probabilistic} by incorporating cosine similarity kernel to model the local manifold structure of documents.
\item \textbf{Successive nonnegative projection (SNPA)} \cite{gillis2014successive}: it is an NMF method. It does not require the matrix $\mathbf{W}$ to be full rank, which makes it more robust to noise than traditional NMF methods.
\item \textbf{Anchor-free correlated topic modeling (AchorFree)} \cite{fu2018anchor}: it is an NMF method. It does not have the anchor-word assumption, which makes it behave much better than traditional NMF methods.
\item \textbf{Deep Poisson Factor Modeling (DPFM)}\cite{henao2015deep}:
    it is a deep learning based topic model built on the Dirichlet process.
    We set its DNN to a depth of two hidden layers, and set the number of the hidden units of the two hidden layers to $c$ and $\lceil c/2\rceil$ respectively for its best performance. We used the output from the first hidden layer for clustering. 
\end{itemize}

We further compared MBN with a cosine-similarity-kernel-based spectral clustering (SC) algorithm \cite{NIPS2001_2092}, and compared DTM with SC+Lasso, for evaluating the effects of MBN on performance.

We evaluated the comparison results in terms of \textit{clustering accuracy} (ACC), \textit{coherence} (Coh.) \cite{neuhaus1998between}, and \textit{similarity count} (SimC.). Coherence evaluates the quality of a single mined topic.
It is calculated by $Coh(\nu) = \sum_{v_1,v_2 \in \nu} \log{\frac{freq(v_1,v_2)+\varepsilon}{freq(v_2)}}$ where $v_1$ and $v_2$ denote two words in the vocabulary, $freq(v_1,v_2)$ denotes the number of the documents where $v_1$ and $v_2$ co-appear, $freq(v_2)$ denotes the number of the documents containing $v_2$, and $\varepsilon = 0.01$ is used to prevent the logarithm operator from zero. The higher the clustering accuracy or coherence score is, the better the topic model is.
Because the coherence measurement does not evaluate the redundancy of a topic, we use the similarity count to measure the similarity between topics.
For each topic, the similarity count is obtained simply by adding up the overlapped words of the topics within the leading $c$ words.
 The lower the similarity count score is, the better the topic model is.

\subsection{Results}


\begin{table}[t]
\setlength{\abovecaptionskip}{0.cm}
\centering
\caption{Comparison results on 20-newsgroups.}\label{tab1}
\scalebox{0.87}{
\begin{tabular}{ccccccc}

  \hline
  \hline
  {Metric}      & {Model}         & {T=5}      & {T=10}     & {T=15}     & {T=20} & {rank} \\ \hline
  \multirow{7}{*}{ACC}
 & {LDA}       & {0.7013} & {0.5915} & {0.5187} & {0.4900} & {5.5} \\
 & {LTM}       & {0.8184} & {0.7109} & {0.6412} & {{0.5996}} & {2} \\
 & {SNPA}      & {0.4078} & {0.3079} & {0.2538} & {0.1744} & {7} \\
 & {AnchorFree} & {0.7595} & {0.6423} & {0.5207} & {0.4485} & {5.25} \\
 & {SC}        & {0.8124} & {0.6960} & {0.5849} & {0.4773} & {3.5} \\
 & {DTM}    & \textbf{{0.8747}} & \textbf{{0.7323}} & \textbf{{0.6471}} & \textbf{0.6538} & {1} \\
 & {DPFM}   & 0.7730 & 0.6439 & 0.5785 & 0.5328 & 3.75 \\
  \hline

 \multirow{7}{*}{Coh.}
 & {LDA}       & \textbf{{-509.76}} & {-574.40} & {-617.87} & {-759.13} & {2} \\
 & {LTM}       & {-893.72} & {-901.88} & {-896.35} & {-855.95} & {6} \\
 & {SNPA}      & {-813.89} & {-843.59} & {-786.52} & {-760.83} & {4.5} \\
 & {AnchorFree} & {-565.95} & \textbf{{-572.25}} & \textbf{{-571.92}} & \textbf{{-596.10}} & {1.5} \\
 & {SC+Lasso}        & {-674.59} & {-762.26} & {-836.73} & {-890.21} & {5.25} \\
 & {DTM}    & {-653.63} & {-728.21} & {-818.98} & {-862.67} & {4.25} \\
 & {DPFM}   & {-1234.22} & {-959.51} & {-696.45} & {-517.01} & 4.5 \\
  \hline

  \multirow{7}{*}{SimC.}
&  {LDA}       & {22.34} & {66.38}  & {116.2}  & {196} & {3.75} \\
&  {LTM}       & {28}    & {30.62}  & {{31.42}}  & \textbf{26} & {2.75} \\
&  {SNPA}      & {31.9}  & {157.02} & {413.48} & {549} & {5} \\
&  {AnchorFree} & {32.7}  & {195.52} & {600.14} & {1235} & {6.25} \\
&  {SC+Lasso}        & \textbf{3.44}  & \textbf{11.68}  & \textbf{24.32}  & {52} & {1.25} \\
&  {DTM}    & {{3.54}}  & {{13.48}}  & {28.04}  & {86.02} & {2.25} \\
&  {DPFM}   & {118.10} & {296.22} & {712} & {890.14} & 6.75 \\
  \hline
  \hline
\end{tabular}}

\end{table}
Table \ref{tab1} shows the comparison results on the 20-newsgroups corpus. From the table, we see that DTM achieves higher clustering accuracy than the other algorithms.
For example, DTM achieves more than $5\%$ absolute clustering accuracy improvement over the runner-up method LTM when $c=5$, and $1\%$ higher in other cases.
In addition, the single-topic quality of the topics mined by DTM ranks the third in terms of coherence. The overlaps between the topics mined by DTM are the smallest in terms of similarity count except the case when $c=15$.

Table \ref{tab2} shows the results on the TDT2 corpus. From the table, we can see that DTM obtains the best performance in terms of all three evaluation metrics. For example, the clustering accuracy produced by DTM is over $3\%$ absolutely higher than that of the runner-up method when mining 5 topics, and over $14\%$ higher than the latter when $c=10$. We further averaged all six ranking lists in Tables \ref{tab1} and  \ref{tab2}.
The average ranking list from the number one to number six is DTM, AnchorFree, SC, LTM, LDA, DPFM and SNPA, respectively.

\begin{table}[t]
\setlength{\abovecaptionskip}{0.cm}
\centering
\caption{Comparison results on TDT2.}\label{tab2}
\scalebox{0.88}{
\begin{tabular}{ccccccc}
  \hline
  \hline
  {Metric}      & {Model}         & {T=5}      & {T=10}     & {T=15}     & {T=20}    & {rank} \\ \hline

  \multirow{7}{*}{ACC}
  &{LDA }      & {0.7013} & {0.6413} & {0.5941} & {0.6093} & {5.75} \\
  &{LTM }      & {0.9443} & {0.7705} & {0.6861} & {0.6458} & {3} \\
  &{SNPA}      & {0.6986} & {0.5612} & {0.4694} & {0.4610} & {7} \\
  &{AnchorFree} & {0.9383} & {0.7756} & {0.7420} & {0.7352} & {2.25} \\
  &{SC}        & {0.7943} & {0.6739} & {0.6266} & {0.5819} & {5.25} \\
  &{DTM}    & \textbf{{0.9778}} & \textbf{{0.9148}} & \textbf{{0.8170}} &\textbf{{0.7842}} & {1} \\
  &{DPFM}   & {0.8037} & {0.7305} & {0.6849} & {0.6776} & 3.75 \\
  \hline

  \multirow{7}{*}{Coh.}
  &{LDA      } & {-509.76} & {-574.40} & {-617.87} & {-642.48} & {4.5 }\\
  &{LTM      } & {-634.29} & {-597.61} & {-579.34} & {-616.12} & {4.25} \\
  &{SNPA     } & {-610.96} & {-668.08} & {-660.27} & {-679.49} & {6 }\\
  &{AnchorFree} & {-407.25} & {-466.23} & \textbf{-494.75} & \textbf{{-531.64}} & {1.5} \\
  &{SC+Lasso}        & {-441.52} & {-517.57} & {-542.88} & {-629.02} & {3.25} \\
  &{DTM}    & \textbf{{-373.89}} & \textbf{{-451.45}} & {{-526.38}} & {{-648.51}} & {2.5} \\
  &{DPFM}   & {-803.90} & {-715.69} & {-676.80} & {-627.00} & 6 \\
  \hline

  \multirow{6}{*}{SimC.}
  &{LDA      } & {8.02  }& {30.48} & {65.08 } & {104.82} & {4} \\
  &{LTM      } & {24.74 }& {23.34} & {23.26 } & {20.76} & {3.5} \\
  &{SNPA     } & {29.36} & {74.78} & {189.44} & {271.5} & {6} \\
  &{AnchorFree} & {6.18}  & {30.42} & {84.18}  & {150.04} & {3.25} \\
  &{SC+Lasso    }    & {1.06}  & {10} & {19.02}  & {35.68} & {2.5} \\
  &{DTM}    & \textbf{{0.3}}  & \textbf{{1.98}}  & \textbf{{5.6}}  &\textbf{{12.32}} & {1} \\
  &{DPFM}   & {112.22} & {287.76} & {690.20} & {1056.20} & 7 \\
  \hline
  \hline
\end{tabular}}

\end{table}

\begin{table}[!t]
\caption{Topics discovered by AnchorFree and DTM}\label{tab3}
\scalebox{1}{
\begin{tabular}
  {|c|c|c|c|c|}
  \multicolumn{5}{c}{AnchorFree}\\
  \hline
  Topic 1 & Topic 2 & Topic 3 & Topic 4 & Topic 5 \\
  \hline

  netanyahu  & \textbf{asian}    & bowl      & tornadoes  & \textbf{economic}       \\
  israeli    & \textbf{asia}     & super     & florida    & {indonesia}    \\
  israel     & \textbf{economic} & broncos   & central    & \textbf{asian}        \\
  palestinian& \textbf{financial}& denver    & storms     & \textbf{financial}    \\
  peace      & \textbf{percent}  & packers   & ripped     & imf          \\
  arafat     & \textbf{economy}  & bay       & victims    & \textbf{economy}      \\
  palestinians& market  & green     & tornado    & \textbf{crisis}       \\
  albright   & stock    & football  & homes      & \textbf{asia}         \\
  benjamin   & \textbf{crisis}   & game      & killed     & monetary     \\
  west       & markets  & san       & people     & {currency}     \\
  \hline
  \multicolumn{5}{c}{DTM}\\
  \hline
  Topic 1 & Topic 2 & Topic 3 & Topic 4 & Topic 5 \\
  \hline
   netanyahu   & asian      & bowl      & florida   & nigeria \\
   israeli     & percent    & super     & tornadoes & abacha  \\
   israel      & indonesia  & broncos   & tornado   & military \\
   palestinian & asia       & denver    & storms    & police \\
   peace       & economy    & packers   & killed    & nigerian \\
   albright    & financial  & green     & victims   & opposition \\
   arafat      & market     & game      & damage    & nigerias \\
   palestinians& stock      & bay       & homes     & anti \\
   talks       & economic   & football  & ripped    & elections \\
   west        & billion    & elway     & nino      & arrested  \\
  \hline

\end{tabular}
}
\end{table}

Table \ref{tab3} shows the top 10 topic words discovered by AnchorFree and DTM respectively when mining a corpus of 5 topics in TDT2. From the table, we can see that DTM produces more discriminative topic words than AnchorFree. Specifically, DTM does not produce overlapping words, while AnchorFree produces 7 overlapping words among the 50 topic words.
The topic words of the second and fifth topics produced by AnchorFree have an overlap of over 70\%. Some informative words discovered by DTM, such as the words related to anti-government activities or violence in the fifth topic, were not detected by AnchorFree. The above phenomenon is observed in other experiments too. We conjecture that the additional conditional assumptions made by AnchorFree, such as consecutive words being persistently drawn from the same topic, might affect the topic characterization. On the contrary, the proposed DTM not only avoids making additional assumptions but also does not suffer the weaknesses of deep neural networks.

\subsection{Effects of hyperparameters on performance}



\begin{figure}[t]
\centering
{\includegraphics[scale = 0.63]{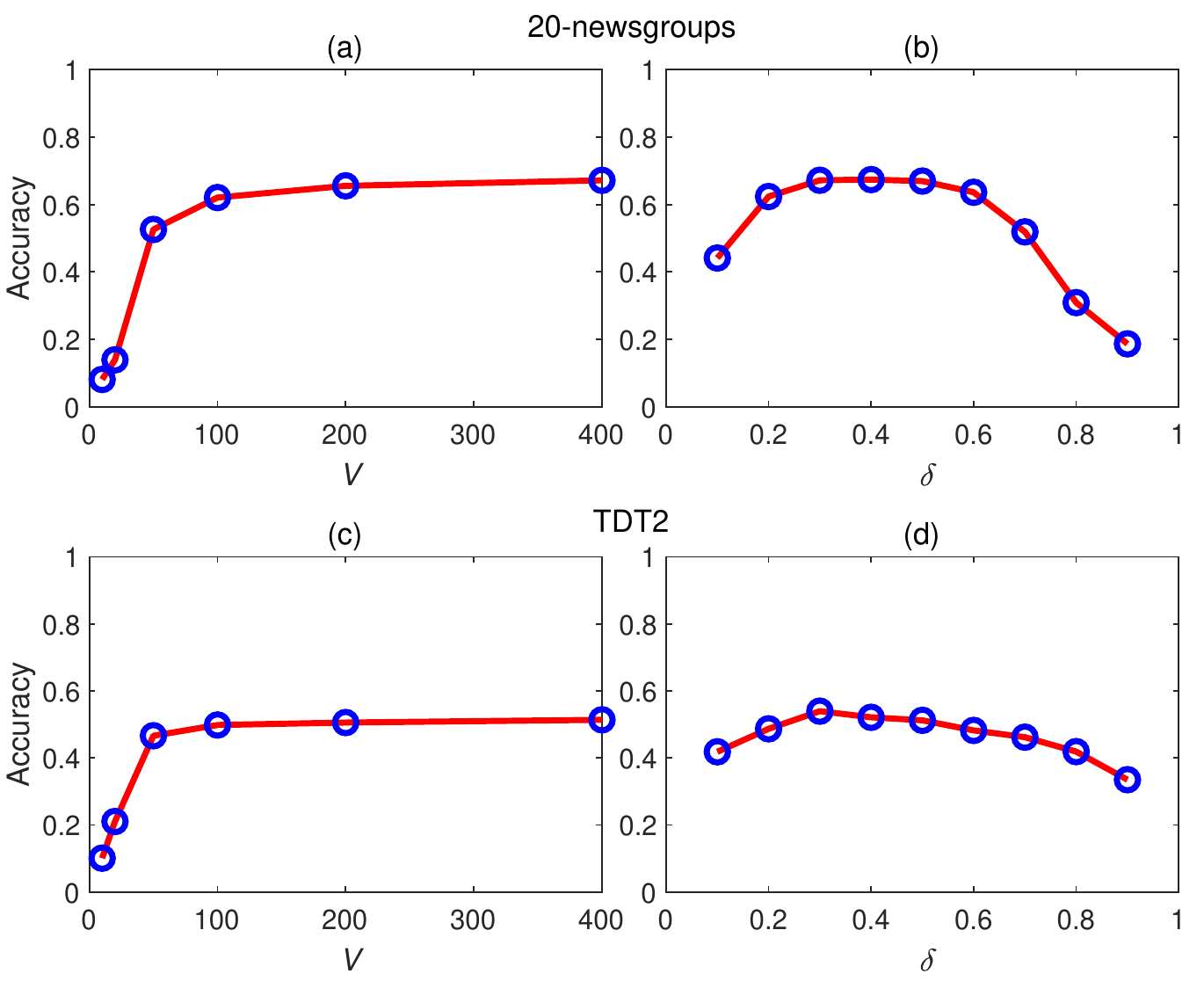}}%
\caption{Effect of hyperparameters $V$ and $\delta$ on performance.}
\label{fig_para}
\end{figure}

We study $V$ and $\delta$ independently on 20-newsgroups and TDT2. When we study a hyperparameter, we tune it in a range, leaving the other hyperparameters to their default values. The experimental results are shown in Fig. \ref{fig_para}. From Figs. \ref{fig_para}a and \ref{fig_para}c, we see that enlarging $V$ increases the accuracy of DTM steadily, and the performance of DTM becomes stable when $V>100$. However, increasing $V$ enlarges the computational complexity of DTM as well. To balance the accuracy and computational complexity, setting $V=400$ is reasonable. From Figs. \ref{fig_para}b and \ref{fig_para}d, we observe that, although the performance is relatively sensitive to hyperparameter $\delta$, the hyperparameter has a stable interval around the default value $0.5$. To conclude, setting $\delta=0.5$ is safe for DTM.


\section{Conclusion}
\label{sec:majhead}
In this paper, we have proposed a deep topic model based on MBN and Lasso. The novelty of DTM lies in the following three respects. First, we extended the linear matrix factorization problem to its nonlinear case. Second, we estimated the topic-document matrix and word-topic matrix separately by MBN and Lasso independently, which simplifies the optimization problem of \eqref{eq:dtm}. At last, we applied MBN and Lasso to the unsupervised topic modeling for the first time. Particularly, MBN, as an unsupervised deep model, overcomes the weaknesses of the model assumptions, anchor word assumption, and shallow learning, which accounts for the advantage of DTM over the 5 representative comparison methods. Experimental results on 20-newsgroups and TDT2 have demonstrated the effectiveness of the proposed method.

\section{Acknowledgement}
We thank Dr.Kejun Huang for sharing the source code of the AnchorFree algorithm.


%



\bibliographystyle{IEEEbib}
\bibliography{strings,refs}

\end{document}